%% file: main.tex
\documentclass[11pt,a4paper]{article}
\usepackage{acl2015}
\usepackage{times}
\usepackage{latexsym}
\usepackage{multirow}

\usepackage{mathtools}
\usepackage{graphicx}
\usepackage{array}

\newcommand{\RNum}[1]{\uppercase\expandafter{\romannumeral #1\relax}}
\newcommand{\TABLE}[1]{Table~\ref{#1}}
\newcommand{\SEC}[1]{Section~\ref{#1}}
\newcommand{\FIG}[1]{Figure~\ref{#1}}
\newcommand{\EQ}[1]{Equation~\ref{#1}}

\author{Ali Basirat and Joakim Nivre\\
  Department of Linguistics and Philology \\
  Uppsala University \\
  {\tt \{ali.basirat,joakim.nivre\}@lingfil.uu.se} \\
}

\title{Greedy Transition-Based Dependency Parsing with{\\}Discrete and Continuous Supertag Features}
\begin{document}
\maketitle
%\tableofcontents
\input{abstract.tex}
\input{introduction.tex}

\input{transition_based_parser.tex}

\input{mica.tex}
\input{feature_templates.tex}
\input{experiments.tex}

\input{conclusion.tex}
\bibliographystyle{acl} 
\bibliography{NLP_Bib}
\end{document}

%% file: abstract.tex
%*\section{Abstract}
%\label{sec:abstract}
\begin{abstract}
We study the effect of rich supertag features in greedy transition-based dependency parsing. While previous studies have shown that 
sparse boolean features representing the 1-best supertag of a word can improve parsing accuracy, we show that we can get further improvements by adding a continuous vector representation of the entire supertag distribution for a word. In this way, we achieve the best results for greedy transition-based parsing with supertag features with $88.6\%$ LAS and $90.9\%$ UAS
on the English Penn Treebank converted to Stanford Dependencies.
\end{abstract}

%% file: introduction.tex
\section{Introduction}
\label{sec:introduction}

%Greedy interesting because
Greedy transition-based dependency parsing is appealing thanks to its efficiency, deriving a parse tree for a sentence in linear time using a feature-based discriminative classifier \cite{yamada2003statistical,nivre2004incrementality}. Although higher accuracy can normally be achieved using beam search and structured 
prediction \cite{zhang08emnlp,huang10,zhang11}, recent research has shown that greedy parsers can be more accurate than traditionally assumed, thanks to techniques 
like dynamic oracles \cite{goldberg12coling,goldberg13tacl}, dynamic parsing strategies \cite{sartorio13}, and neural network classifiers using dense continuous feature representations
\cite{chen14}. Another recent line of research has addressed the need for more informative features. Hence, both \newcite{ambati14} and \newcite{ouchi14} have shown that
using a supertagger, in addition to a traditional part-of-speech tagger, can bring significant improvements to a greedy dependency parser.

In this paper, we continue to explore the use of supertag features in greedy transition-based dependency parsing. We use MaltParser with SVM classifiers and 
a standard feature model as our baseline system, and extend it with features defined over the supertags used by the MICA parser \cite{Bng09}. 
The supertags are elementary trees in a Tree-Insertion Grammar (TIG) \cite{Schabez_TIG} that have been automatically extracted from the Wall Street Journal
 part of the Penn Treebank \cite{Marcus93_WSJ} using the approach of \newcite{Che01}. The TIG elementary trees encapsulate syntactic environments in 
 which a word can appear. 
 
 We first demonstrate that 1-best supertags as features over words improve parsing accuracy, thus reproducing the results of  \newcite{ambati14} and \newcite{ouchi14} in a novel setting. We then go on to show that additional improvements can be obtained by adding continuous
 features representing the complete probability distribution of supertags for a word, mapped to a lower dimensionality using PCA. This result is interesting as it shows 
 the potential value of continuous features not just in a neural network setting but also using conventional linear classifiers.

%% file: transition_based_parser.tex
\section{Transition-Based Dependency Parsing}
\label{sec:trans_based_dep_parsing}

A greedy transition-based dependency parser derives a parse tree for a sentence by predicting a sequence of transitions between parser configurations. 
The parsing process starts from an initial configuration and ends with some terminal configuration. 
A configuration is characterized by a triple $c=(\Sigma,B,A)$, where $\Sigma$ is a stack that stores partially processed words, $B$ is a buffer that stores unprocessed words in the input sentence, and $A$ is a parse tree assigned to the processed words. 
%$Nodes$ are positive integers corresponding to linear position of the words in the input sentence and one extra artificial root node $0$.

Transitions between configurations are controlled by a classifier that usually takes the form of a discriminative model such as an SVM or a log-linear model.
The classifier uses a history-based feature model that combines features of the partially built dependency tree, representing the derivation history, and 
attributes of the input sentence. 
%\shortcite{chen2014fast} showed that lexical features are indispensable for making accurate decision in transition-based dependency parsing models. 
%However they are incomplete and very expensive to compute. 

Different parsing algorithms have been proposed for moving between configurations. In this paper, we use the arc-standard algorithm, also known in 
the MaltParser implementation as stack projective \cite{nivre04acl,nivre09acl}. 
The algorithm starts in an initial configuration where all words of the sentence are in the buffer and a dummy root word is in the stack.
It uses the three actions \emph{Shift}, \emph{Right-Arc}, and \emph{Left-Arc} to transition between the configurations and build the parse tree.  
It ends in a terminal configuration where the buffer is empty and the stack again contains only the dummy root word.
%The two actions Left-Arc and Right-Arc are used to build left and right dependencies respectively.
%These actions are restricted to the fact that the final dependency tree has to be rooted at node $0$. 

Using $s_i$ to denote the $i$-th element in the stack and $b_j$ for the $j$-th element in the buffer, the actions are defined as follows:
\begin{itemize}
\item \emph{Shift} pushes $b_0$ onto the stack.% unconditionally.  
%It enables algorithm to proceed with words in the input sentence from left to right. 
\item \emph{Right-Arc} makes $s_0$ a right dependent of $s_1$ and removes $s_0$ from the stack. 
%This action is restricted to the case that the next token in the buffer is a dependent of the root node $0$ in the partially constructed dependency tree. 
\item Left-Arc makes $s_1$ a left dependent of $s_0$ and removes $s_1$ from the stack. 
%This action is restricted to the case that the top element in the stack is a dependent of root node $0$ in the partially constructed dependency. 
%Moving $s_1$ back to the buffer, Swap change the order of the nodes so that the non-projective trees can be constructed. 
%The Swap action is allowed only once for each node. 
\end{itemize}
The decision between possible actions in each configuration is made by the classifier based on the features which usually describe prefix 
nodes in $\Sigma$ and $B$, and the relationships between these nods and certain nodes in partially built tree $A$. 
In \SEC{sec:feature_templates}, we describe the feature templates used in our experiments. 

%The transition-based dependency parsing follows the data-driven approaches of parsing, in which a parser learns to analyse a sentence based on examples of correctly parsed data. 

%MaltParser is characterized as a parser generator tool for inducing a data-driven dependency parser from a treebank. 
%It is an implementation of inductive dependency parsing in which a global parse tree for a sentence is constructed from a series of local parse trees associated with parts of the sentence \cite{}. 

%% file: mica.tex
\section{MICA Supertags}
\label{sec:MICA}

MICA \cite{Bng09} is a dependency parser that returns deep dependency representations of a given sentence with an accuracy of $87.6\%$ for unlabeled dependency trees and an accuracy of $85.8\%$ for labeled dependencies on section 00 of the Wall Street Journal section of the Penn Treebank. 
It uses two grammars for computing the $n$-best parse trees of an input sentence, a Tree Insertion Grammar (TIG) and a Probabilistic Context Free Grammar (PCFG). 
The former grammar, here known as the \emph{MICA grammar}, contains $4,726$ tree frames associated with about one million words. 
The latter grammar is a large-scale PCFG that generates strings of elementary trees. 
This grammar is directly obtained from the TIG by doing systematic transformations on its elementary trees, as described in 
\newcite{Bng09}. 

Parsing in MICA is carried out in two steps: supertagging and actual parsing.
In supertagging \cite{Bng99}, it uses a Maximum Entropy model to assign elementary trees of the TIG to the words of an input sentence.
The accuracy of supertagging in MICA is $88.52\%$.  
In the actual parsing, it then builds a PCFG from the elementary trees assigned to the input words and derives a set of parse trees from the PCFG rules. 
In the experiments reported in this paper, we only make use of the MICA supertagger.
%The SYNTAX \cite{Bou88} parser generator is used to generate an Earley-like parser for analyzing the PCFG. 
%The Earley-like parser computes a shared parse forest containing all parse trees of the input sentence. 
%Then, an $n$-best filtering module is used to filter the forest. 
%The result is an $n$-best forest of constituency and dependency parse trees with a clear boundary between the best parse trees and the others. 

%% file: feature_templates.tex
\section{Feature Templates}
\label{sec:feature_templates}

We now describe the feature templatess used in our experiments, starting with the features of the baseline model
and continuing with our two different kinds of supertag features.
%This section details different feature templates we have used in our experiments. 
The following notation is used to describe the features: 
\begin{itemize}
\item The symbols $\Sigma_i$ and $B_i$ refer to the $i$-th word from the top of the stack, and the $i$-th word in the buffer, respectively. 
\item The symbols $w$ and $t$ denote the word form and POS tag of a word, respectively.  
\item The symbols $ld$ and $rd$ denote the leftmost and rightmost dependents of a word, respectively; $h$ denotes the head of the word
and $r$ the dependency relation to the head. 
\item The operator $:$ is used to conjoin features. 
\end{itemize}

\subsection{Baseline Features}
Our baseline model (BL) is the pre-trained MaltParser model for English, available on the MaltParser website and evaluated in \newcite{nivre10coling}:
\begin{itemize} 
  \item Single-word features: $\Sigma_0.w$, $\Sigma_1.w$, $\Sigma_2.w$, $B_0.w$, $B_1.w$, $\Sigma_0.ld.w$, $\Sigma_0.ld.t$, $\Sigma_0.rd.t$, $\Sigma_1.ld.t$, $\Sigma_1.rd.t$, $\Sigma_0.ld.r$, $\Sigma_0.rd.r$, $\Sigma_0.rd.w$, $\Sigma_0.t$, $\Sigma_1.t$, $\Sigma_2.t$, $\Sigma_3.t$, $B_0.t$, $B_1.t$, $B_2.t$
  \item Two-word features: $\Sigma_0.t  \! : \!  \Sigma_1.t$, $\Sigma_0.w \! : \! B_0.w$, $\Sigma_0.t \! : \! \Sigma_0.w$, $\Sigma_1.t \! : \! \Sigma_1.w$, $B_0.t \! : \! B_0.w$, $\Sigma_1.rd.r \! : \! \Sigma_0.ld.r$
  \item Three-word features: $\Sigma_0.t \! : \! \Sigma_1.t \! : \! B_0.t$, $\Sigma_0.t \! : \! \Sigma_1.t \! : \! \Sigma_2.t$, $\Sigma_0.t \! : \! B_0.t \! : \! B_1.t$, $B_0.t \! : \! B_1.t \! : \! B_2.t$, $B_1.t \! : \! B_2.t \! : \! B_3.t$, $\Sigma_1.rd.t \! : \! \Sigma_1.ld.t \! : \! \Sigma_1.t$, $\Sigma_1.t \! : \! \Sigma_1.ld.r \! : \! \Sigma_1.rd.r$
\end{itemize}
Each feature template is internally converted to a sparse vector of boolean features.

\subsection{MICA Supertag Features} 
The output of the MICA supertagger for each word in an input sentence is a probability distribution over the set of supertags given the word. 
We have designed two feature models to exploit this information. %: 
%\begin{enumerate}
%  \item best supertag
%  \item combined-supertag
%\end{enumerate}
The first model, called \emph{best-supertag} (BS), only includes features defined over the most probable supertag assigned to a word. 
Denoting the most probable supertag assigned to a word by $bs$, the best-supertag model includes the following feature templates:  
\begin{itemize} 
  \item single-word features: $\Sigma_0.bs$, $\Sigma_1.bs$, $\Sigma_2.bs$, $\Sigma_3.bs$, $B_0.bs$, $B_1.bs$, $B_2.bs$, $B_3.bs$
  \item two-word features: $\Sigma_0.bs \! : \! \Sigma_1.bs$, $\Sigma_0.bs \! : \! \Sigma_0.w$, $\Sigma_1.bs \! : \! \Sigma_1.w$, $B_0.bs \! : \! B_0.w$
  \item three-word features: $\Sigma_0.bs \! : \! \Sigma_1.bs \! : \! B_0.bs$, $\Sigma_0.bs \! : \! \Sigma_1.bs \! : \! \Sigma_2.bs$, $\Sigma_0.bs \! : \! B_0.bs \! : \! B_1.bs$, $B_1.bs \! : \! B_2.bs \! : \! B_3.bs$
\end{itemize}
Each of these feature templates is again converted to a sparse vector of boolean features.

The second model, called \emph{supertag-distribution} (SD), includes information about the probability of \emph{all} supertags for a word. 
This model relies on the list of probabilities $p(s_i \mid w)\ i=1\dots n$, where $s_i$ is the $i$-th supertag in the MICA grammar containing $n$ supertags ($4726$), and $w$ is an input word.  
The list of these probabilities for each word can be viewed as an $n$-dimensional vector, called supertag vector, in a vector space whose dimensions correspond to supertags in the MICA grammar. The component for each dimension is a real number between $0$ and $1$. 
Given the vector space, each word in a sentence can be represented by an $n$-dimensional vector. 
%Denoting each probability by $P(m\Sigma_i \mid w)$, where $m\Sigma_i$ is a supertag in the MICA grammar and $w$ is a word, as $p_i$, the full supertag distribution model is defined as $\cup_{i=1}^n \Sigma_0.p_i, \Sigma_1.p_i$.
Since these vectors are numerical, they can be used directly as real-valued features in a linear classifier. 

However, preliminary experiments showed that the large number of dimensions and the sparsity in the supertag vectors can lead to low parsing speed.  
%Our experiments on the training data show that on average $y$ elements of these vectors are zero values. %, which leads to a drastic reduction in parsing speed and training. 
Therefore, instead of using the high-dimensional supertag vectors directly, the SD model makes use of vectors obtained by projecting the supertag vector to a lower dimensional vector space using Principal Component Analysis (PCA). 
%
%PCA is a linear algebraic tool which is widely used for representing noisy data in a new basis with less or equal dimensions. 
%It uses a set of variables, called \emph{principal components}, to transform a vector in the original space to the new space. 
%
In this method, each word in a sentence is represented as a vector in a $k$-dimensional vector space ($k \leq n$), where dimensions correspond to the $k$ first principal components of training data. 
Given the principal component matrix $P_{n\times k}$, a vector $X_w$ in the $n$-dimensional supertag vector space corresponding to a word $w$ can be converted to the vector $Y_w$ in the $k$-dimensional vector space using \EQ{eq:pca_transformation}.
%Each word $w$ in this vector space is represented by a $k$ dimensional vector $Y_w$ as below: 
\begin{equation}
  Y_w = P^T X_w
\label{eq:pca_transformation}
\end{equation}
The SD feature model includes all elements of the vector $Y_w$ as distinct continuous features. 
%So, number of features in the reduced supertag feature template is a function of the number of dimensions $k$ in the PCA transformation. 
Denoting each element of $Y$ by $y_i$, the feature templates used can be defined as follows:
\begin{equation}
 \cup_{i=1}^k \Sigma_0.y_i, \Sigma_1.y_i
 \end{equation}
 In other words, we include the $k$-dimensional feature vectors only for the word on top of the stack and the first word in the buffer.

%% file: experiments.tex
\section{Experiments}
\label{sec:experiments}

Our experiments have been run on the WSJ section of the Penn Treebank \cite{Marcus93_WSJ}, using sections 02-21 as training set, section 22 as development set, and section 23 as test set. The Stanford dependency conversion tool was used for converting the WSJ phrase structure trees to basic Stanford 
dependencies \cite{de2006generating}. We used MXPost \cite{adwait96} for POS tagging, with ten-way jackknifing on the training set, and we used MICA \cite{Bng09}
for supertagging. All the dependency parsing models were trained using MaltParser \cite{nivre06malt} with a liblinear multiclass SVM and the stack projective parsing
algorithm.

\begin{figure}[t]
  \begin{center}
    \includegraphics[width=7.5cm]{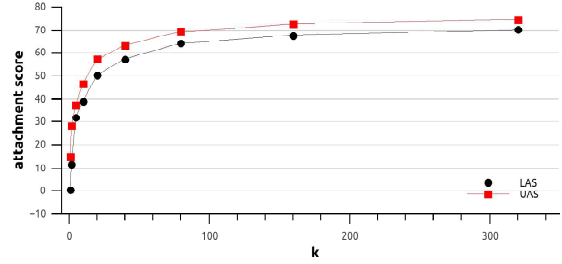}
  \caption{The impact of number of principal components ($k$) on parsing accuracy (UAS, LAS); development set.}
  \label{fig:k_pca}
  \end{center}
\end{figure}

\subsection{Tuning the SD Model}

The dimensionality $k$ of the PCA reduction is a hyper-parameter of the SD model. \FIG{fig:k_pca} shows the accuracy on the development
set when varying the number $k$ of principal components. We see that increasing $k$ has
a positive effect on parsing accuracy, but that the improvement seems to level out around 300. A labeled attachment score of about 70\% may 
not seem very impressive, but it must be remembered that the pure SD model does not incorporate any POS features or lexical features.
In addition, it is restricted to just the two words $\Sigma_0$ and $\Sigma_1$. 

\begin{table}[t]
\begin{center}
\begin{tabular}{|l|r|r|}
\hline
\textbf{Features}        &   \textbf{UAS}   &   \textbf{LAS}      \\ \hline
FORM    &   56.55 &   51.57    \\        
POS     &   49.00 &   41.90    \\        
SUPERTAG&   74.56 &   68.81    \\
SD      &   74.65 &   70.18    \\ \hline
\end{tabular}
\caption{Accuracy of different parsing models built on the two top words of the stack (UAS, LAS); development set.}
\label{table:two_top_nodes}
\end{center}
\end{table}

For comparison, Table~\ref{table:two_top_nodes} shows the accuracy achieved when restricting features to the two top words on the stack, 
but using different types of information. We see that supertag features (SUPERTAG, SD) are vastly superior to POS tags (POS) as well as word forms (FORM),
and we see that the continuous features in the SD model (with $k = 320$) do slightly better than the boolean features in the SUPERTAG model when it comes to labeled
attachment score. 

\begin{table}[t]
\begin{center}
\begin{tabular}{|l|r|r|}
\hline
\textbf{Features}        &   \textbf{UAS}   &   \textbf{LAS}      \\ \hline
BL            &   90.29 &   87.72    \\        
%SD            &   89.88 &   86.14    \\ 
%\hline \hline
BL+BS         &   91.75 &   89.80    \\        
BL+SD         &   91.34 &   89.41    \\         
%BL+SD         &   91.58 &   89.67    \\ 
%\hline \hline
BL+BS+SD      &   91.89 &   89.99    \\
%BL+BS+SD      &   91.93 &   89.97    \\
%BL+CS+SD      &   91.99 &   90.08    \\ \hline \hline
%BL+BS+CS+SD   &   92.13 &   90.18    \\ 
\hline
\end{tabular}
\caption{Accuracy of different parsing models on the development set; BL = baseline, BS = best supertag, SD = supertag distribution.}
\label{table:acc_devset}
\end{center}
\end{table}

\subsection{Combined Feature Models}

\TABLE{table:acc_devset} shows the accuracy of different feature model combinations on the development set. The SD model in this case is trained with $k=320$ principal components. We see that adding supertag features to the baseline model improves parsing accuracy significantly, regardless of whether we use the BS or the SD model. When used by themselves, the BS model gives a slightly larger improvement than the SD model, but a combination of the two models is better than any of the models by themselves. This indicates that the distributional supertag model captures partly different information from the best supertag model.

When evaluated on the final test set, the combined model BL+BS+SD achieves 90.92\% UAS and 88.62\% LAS, which is an improvement by 0.66 (UAS) and 0.73 (LAS) percent absolute compared to the baseline model. We can compare this with \newcite{ambati14}, who report 90.56\% UAS (+0.24) and 88.16\% LAS (+0.29) when using MaltParser with CCG supertags.\footnote{The results of \newcite{ouchi14} are not directly comparable, as they use a different dependency conversion of the WSJ data.}
The results indicate that MICA supertags are at least as effective as CCG supertags and that we seem to get an additional improvement by combining discrete and continuous supertag features.

%% file: conclusion.tex
\section{Conclusion}
\label{sec:conclusion}

Supertags provide a rich syntactic concept that can incorporate the global syntactic environment of a word into a local representation. 
We have studied the effect of using supertag features derived from the MICA supertagger on the accuracy of parsing with a greedy transition-based dependency parser. 
We have corroborated earlier results from \newcite{ambati14} and \newcite{ouchi14} by demonstrating that symbolic (or binary) features defined over the 
1-best supertag of a word has a positive impact on parsing accuray. In addition, we have shown that using continuous features representing the entire supertag 
distribution for a word, with suitable dimensionality reduction, can have an equally positive effect. And combining the two feature types leads to additional improvements.

An interesting line of future work is to investigate how the use of supertag features interacts with orthogonal approaches to improving the accuracy of greedy
transition-based dependency parsers, including the use of dynamic training oracles \cite{goldberg12coling,goldberg13tacl} and dynamic parsing strategies
\cite{sartorio13}. The results obtained with the SD model also suggest that the use of continuous features, recently exploited as latent representations in 
neural network classifiers, may be underexploited in the more traditional approach based on linear classifiers.

%An interesting line of future work is to test the effect of other supertag representations (\eg CCG, HPCG) on the accuracy of transition-based dependency parsing. 
%It could also be interesting to see how the supertag-features can affect the parsing approaches that embed the features in a neural network.